\documentclass{article}
\usepackage{acra}

\usepackage{subfig} 
\usepackage{epsfig} 
\usepackage{epstopdf}
\usepackage{bm}
\usepackage{amssymb} 
\usepackage{amsmath} 

\usepackage{algorithm} 
\usepackage{algpseudocode} 
\usepackage{amsmath, balance} 

\title{Iterative Smoothing and Outlier Detection for \\Underwater Navigation}
\author{Sajad Hassan, Hongkyoon Byun and Jonghyuk Kim\\ Robotics Institute \\ University of Technology Sydney   \\
\{sajad.hassan@student, hongkyoon.byun@student, jonghyuk.kim@\}uts.edu.au
}

\begin{document}

\maketitle

\begin{abstract}

 Underwater visual-inertial navigation is challenging due to the poor visibility and presence of outliers in underwater environments. The navigation performance is closely related to outlier detection and elimination. Existing methods assume the inertial odometry is accurate enough for outlier detection, which is not valid for low-cost inertial applications. We propose a novel iterative smoothing and outlier detection method aiming for underwater navigation. Using the dataset collected from an underwater robot and fiducial markers, experimental results confirm that the method can successfully eliminate the outliers and enhance navigation accuracy. 

\end{abstract}


\section{Introduction} \label{sec:Introduction}

There has been an increasing interest in accurate underwater navigation for underwater inspection and intervention applications.  Intervention is becoming one fundamental task for underwater industrial applications: construction or maintenance of underwater infrastructures \cite{Bonin-FontFrancisco2015Vsfa} \cite{ChavezArturoGomez2019Unuv}. The dynamic and hydrodynamic modelling of the vehicle requires accurate navigation information. Particularly when the vehicle is close to the object to be manipulated \cite{NamDinhVan2020RSVI}. Integrated inertial navigation systems aided by vision or acoustic sensors have been the primary approach for underwater navigation.

\begin{figure}[t]\centering
\epsfig{figure=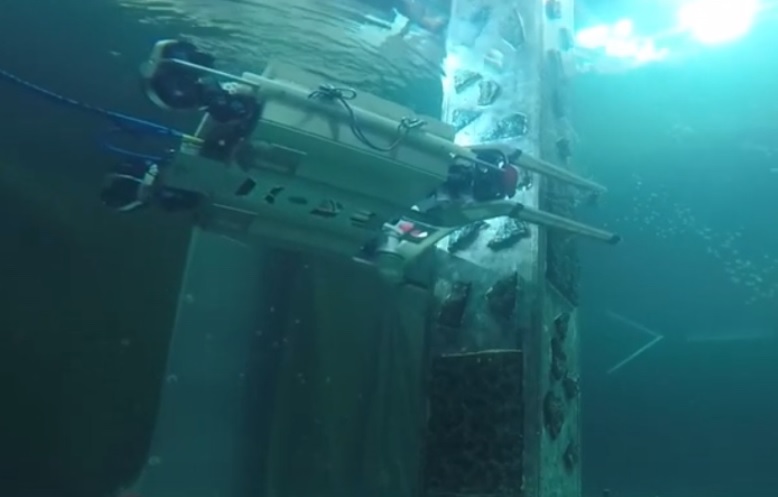,width=1\linewidth}
\caption{The Submersible Pile Inspection Robot (SPIR) developed at the UTS performing underwater inspection and maintenance tasks in a test water-tank facility.} \label{fig:1}
\end{figure}
    
\par While the Kalman filter is a valuable tool in robotics and navigation applications, its performance is severely impaired by outliers in the measurements. Optimal smoothing state estimation can also be degraded by the presence of outliers in the measurements \cite{KautzThomas2015ArKf}. According to \cite{alma991032133873605106} "an outlier is an observation which deviates so much from the other observations as to arouse suspicions that it was generated by a different mechanism.” Outliers may occur due to environmental disturbances, noisy sensor measurement or temporary sensor failures \cite{AgamennoniGabriel2011AoKf}. For example, computer vision data contaminated by outliers or sonar data corrupted by phase noise result in erroneous measurements \cite{LeeKyuman2020ROFf}. 

Optimal smoothing further improves state estimation in the Kalman filter. Optimal smoothing can utilise the measurements made after the current estimation time \cite{GrewalMohinderS2014OS}. In this work, we propose a new iterative smoothing technique to detect and remove outlier measurements for more accurate robot state estimation, which is required for precision underwater interventional tasks as exampled in Figure \ref{fig:1}. In particular, we utilise Biswas-Mahalanabis Fixed-lag smoother (BMFLS) to detect outliers at the fixed-lag time step from when the data is received. This method is quite computationally efficient and stable compared to other techniques such as EM (Expectation and Maximisation) or two-pass smoothers.
    
Considering the low-quality inertial odometry and its fast drifting solution, we treat all visual measurements as inliers in the first smoothing iteration. The outliers are declared using the standard chi-squared ($\chi^2$) gating method. The trajectory is then refined without using the outliers. By repeating these steps, the estimated trajectory and set of outliers can converge.

Our key contributions are as follow:
\begin{itemize}
\item A novel iterative smoothing and outlier detection algorithm, utilising Biswas-Mahalanabis Fixed-lag Smoother (BMFLS). It addresses the problem of rapid drift in the inertial odometry and iteratively classifies the outlier candidates.
\item We demonstrate the method using a dataset collected from an underwater robot and fiducial (ARTag) markers. The dataset suffers from noisy, missing measurements due to the confusion in recognising the markers, thus resulting in outliers.
\end{itemize}

The remainder of the article is outlined as follows. Section~\ref{sec:Related} discusses the literature review of the work done in outlier detection. Section \ref{sec:SystemModels} details the iterative Biswas-Mahalanabis Fixed-lag Smoother with outlier detection algorithm. Section~\ref{sec:experiments} will provide experimental results using data collected from a water-tank environment. Section~\ref{sec:conclusions} will conclude with future direction.

\section{Related Work} \label{sec:Related}


Many approaches have been studied to make the Kalman filter and its variants more robust towards measurements with outliers. Some of these methods require manual filter-tuning or low-dimensional, and it cannot be extended well into higher-dimensional problems \cite{AgamennoniGabriel2011AoKf}. Identifying outliers in multivariate data also poses challenges compared to univariate data \cite{GhorbaniHamid2019MDAI}.

\cite{AgamennoniGabriel2011AoKf} introduced the outlier-robust Kalman filter (ORKF) that uses the Student-$t$ distribution to model the outliers with a heavy-tail distribution which is not well described in the Gaussian distribution. A fixed-interval smoother is used, which is not well suited for near-realtime estimation. Some work is simulation-based (no real data), and outliers are artificially added to the observations \cite{ChangGuobin2014RKfb}. 

While \cite{ChangGuobin2014RKfb} introduced a robust Kalman filter based on Mahalanobis distance (MD) as judging criteria for outlier detection. Suppose the variance of the current innovation is greater than the Chi-square distribution threshold. In that case, a scaling factor is introduced to rescale the covariance of the observation noise to reduce the filter Kalman gain to maintain robustness. However, this method may not be suitable for outlier detection for a real-time application and may lead to sub-optimal performance.
    
The chi-square test is also applied for fault detection in the navigation field\cite{WangRong2016CaSc}.\cite{LeeKyuman2020ROFf} combines the random sample consensus (RANSAC) and chi-square test using the Mahalanobis gating test along with the extended version of the ORKF to detect and handle measurement outliers in vision-aided estimation problem. They have initially used RANSAC to provide clean data to the filter. For the remaining outliers that are not detected in the image processing, the Mahalanobis gating test is used.
    
Further, the MD is also used by \cite {KautzThomas2015ArKf} as a threshold to detect outliers. The authors introduce a robust Kalman filter with re-sampling and optimal smoothing.  In their work, if an observation is greater than the defined threshold, it will be rejected from the fixed-lag smoother.

Furthermore, \cite{HadiAliS1992IMOi} introduced an iterative MD approach for detection of outliers in multivariate data that also may be affected by masking or swamping issues. However, this method may not be feasible for Kalman filter applications.

In work mentioned above, the critical assumption is that the odometry solutions are reasonably accurate for outliner detection. If low-cost inertial sensors are used, this assumption is not valid anymore. The direct and catastrophic consequence is that the inertial errors can reject all good measurements. For example, \cite{ChavezArturoGomez2019Unuv} admits localisation is a challenging task in underwater environment in their work. They have proposed to use aritifical markers to improve the navigation in their intervention missions where the obvious outliers are from sensor readings are rejected heuristically.
Also, few of the work mentioned have taken the advantage of optimal smoothing in outlier detection problem.
We address this problem by introducing an iterative smoothing and detection and including all measurements as inliers in the first iteration.

\section{Iterative Smoothing and Outlier Rejection}
\label{sec:SystemModels}

A Biswas-Mahalanobis fixed-lag smoother is used to provide the smoothed state. BMFLS augments the state vector $x_{k[s]} = \{x_k, x_{k-1},\cdots,x_{k-N}\}$ using $N$-lagged states and estimate the $x_{k-N}$ state, in which $k$ is the current measurement time. The augmented state is use to run similar to an extended Kalman filter (EKF), yielding an computationally efficient performance.

The state dynamic model is nonlinear consisting of inertial navigation equations with $3$D coordinate transformation. The state model included the position and velocity in $x,y,z$ directions, and Euler angles $(\phi, \theta, \psi)$. The measurement model is linear in this case as the visual sensor delivers direct pose measurments. That is,
 \begin{align} 
  {x}_{k} &= f(x_{k-1})+g(w_k), \qquad w_k \sim N(0,Q_k) \\
  {z}_k &= H(x_k)+v_k, \qquad v_k\sim N(0,R_k),
  \end{align} 
  where the measurement matrix $H$, process and measurement noise covariance matrices $Q$ and $E$ are constructed as follows:
  \begin{align}
 H&= \begin{bmatrix}
    1& 0& 0& 0& 0& 0& 0& 0& 0\\0& 1& 0& 0& 0& 0& 0& 0& 0\\
    0& 0& 1& 0& 0& 0& 0& 0& 0\\
    0& 0& 0& 0& 0& 0& 1& 0& 0\\
    0& 0& 0& 0& 0& 0& 0& 1& 0\\
    0& 0& 0& 0& 0& 0& 0& 0& 1
    \end{bmatrix}\\      Q&=diag[\sigma^2_x,\sigma^2_y,\sigma^2_z,\sigma^2_\phi, \sigma^2_\theta,\sigma^2_\psi]\\
      R&=diag[\sigma^2_x,\sigma^2_y,\sigma^2_z,\sigma^2_\phi, \sigma^2_\theta,\sigma^2_\psi]\\
      G &= \begin{bmatrix}
      0& 0& 0& 0& 0& 0\\0& 0& 0& 0& 0& 0\\
      0& 0& 0& 0& 0& 0\\
      & & & 0& 0& 0\\
      & C^n_b& & 0& 0& 0\\
      & & & 0& 0& 0\\
      0& 0& 0&  \\
      0& 0& 0& & C^n_b\\
      0& 0& 0
      \end{bmatrix}
  \end{align}

 The IMU (inertial measurement unit) outputs are the body acceleration $f^b$ and angular velocity $\omega^b$. Euler angle integration is applied for the orientation prediction. The state prediction model is as follows:
    \begin{align}
        \begin{bmatrix}
        \hat{p}_{k} \\ \hat{v}_{k} \\ \hat{\Psi}_{k} \\
        \end{bmatrix}_{9\times 1} &= \begin{bmatrix}
        \hat{p}_{k-1}+\hat{v}_{k-1}\Delta t \\ \hat{v}_{k-1}+(C^n_b f^b_k+g)\Delta t \\ \hat{\Psi}_{k-1}+(E^n_b\omega^b_k)\Delta t
        \end{bmatrix}, \label{eq:nonlinear}
    \end{align}
where $C^n_b$ represents the body to frame direction cosine matrix, and $E^n_b$ represents the body rotation to Euler angular rates transformation matrix, $g$ represents the gravitational acceleration and $\Delta t$ denotes the time increment.
    
The BMFLS state is propagated using the nonlinear dynamic model and linearised Jacobian matrix, 
\begin{align} \label{K_FLS}
     \hat{{x}}_{k+1[s]} &= f(\hat{{x}}_{k[s]}) \\
\hat {{P}}_{k+1[s]}&=\Phi_{k[s]}\hat{P}_{k[s]}\Phi^{T}_{k[s]}+G_{k}{Q}_{k[s]}G^T_{k}\Delta t, 
\end{align}
where $\Phi_{k[s]}$ is the linearised and discrete state transition matrix which is construcnted from the Jacobian of the Equation \ref{eq:nonlinear} and fixed-lag sliding-window transition of the states. 

When the vision measurements are available, an innovation (or error) and its covariance is computed,
\begin{align}
       e_{k[s]} &= c(z_k-H\hat{x}_{k[s]}) \\
       S_{k[s]} &= {H}P_{k[s]}H^T+R,
\end{align}
where $c \in\{0,1\}$ is the outlier association variable. When the measurement is an inlier, $c=1$, otherwise $c=0$.
\begin{align}
       d_k &= e^T_{k[s]}S_{s[k]}e_{k[s]} \\
       c &= \left\{ \begin{array}{c} 1, \; \text{if} \;d_k < \chi^2_T \\ 0, \; \text{if} \;d_k \geq \chi^2_{T}\end{array}, \right.
\end{align}
where $\chi^2_{T}$ is the threshold value selected from the degrees-of-the freedom (DOF) of the measurement.

The BMFLS state and covariance are then updated using the smoothing Kalman gain,
\begin{align}
     K_{k[s]} &= P_{k[s]}H^T S^{-1}_{k[s]} \\ 
        \hat{{x}}_{k[s]}&=\hat{{x}}_{k[s]}+ {K}_{k[s]}e_{k[s]} \\
        {P}_{k[s]}&={P}_{k[s]}-{K}_{k[s]}{H}{P}_{k[s]}.
\end{align}

Algorithm 1 illustrates the iterative smoother outlier detection process. It works by treating all the measurements in the dataset as inliers initially. In the first iteration, the first set of outliers are detected and removed from the measurements. In the 2nd iteration, the dataset is run without the first outlier set captured in the 1st iteration. The 2nd outlier second from this iteration is also removed. This process is repeated until the number of inliers converges.
    
    \begin{algorithm}[t]
        \caption{Iterative Smoothing and Outlier Detection}
        \label{Alg_1}
        
            	\begin{algorithmic}[1]
            	\State Initialisation: \(x_0\), \(P_0\), Q, R, \(\chi^2_{threshold}\)
            	\While{No of inliers converge}
            	    \For {$k=1:T$}
            	    
            	    \State \(x_s\) =BMFLS Function()

            	    \If {measurement available}
            	   \State \(e_{k[s]}\)=\(z_k-Hx_{k-[s]}\)
            	    \State \(S_{s[k]}=HP_{k-[s]}H^T+R\)
            	    \State
            	    \(d_k\) = \(e^T_{k[s]}S_{s[k]}e_{k[s]}\)
            	    
            	    \State perform outlier gating test:
            	    \If {$d_k>$$\chi^2_{threshold}$}
            	    \State {store as outlier}
            	   \Else
            	   \State store as inlier
            	   
            	    \EndIf
            	    \EndIf

            	    \EndFor
            	    \State {plot(states)}
            	 \State remove outliers
            	 \EndWhile
            	 
            	\end{algorithmic}

    \end{algorithm}

\section{Experimental Results}\label{sec:experiments}

\begin{figure}[t]\centering
\epsfig{figure=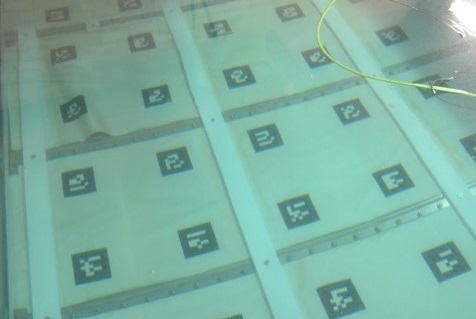,width=1\linewidth}
\caption{ARTag-based fiducial markers installed on the floor of the water tank, which is used to provide the pose (position and orientation) measurements using a monocular camera installed in the robot. The markers' positions and orientation are pre-calibrated.} \label{fig:2}
\end{figure}

\begin{figure}[t]\centering
\epsfig{figure=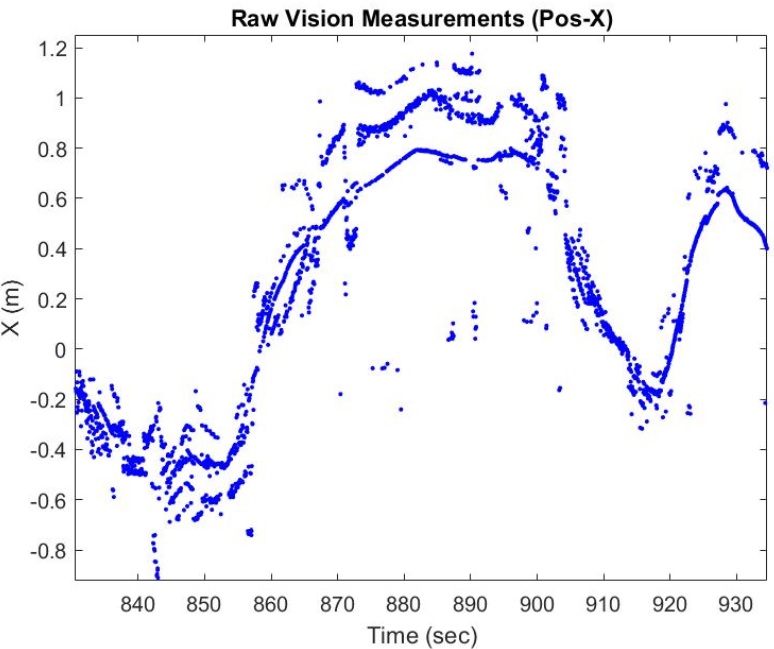,width=1\linewidth}
\caption{An example of outliers from the raw vision and markers measurements. It can be seen that the noises do not follow the standard Gaussian statistics but an offset-like outlier pattern, which stems from the confusion in recognising the markers.} \label{fig:3}
\end{figure}

\begin{figure}[th!]\centering
\epsfig{figure=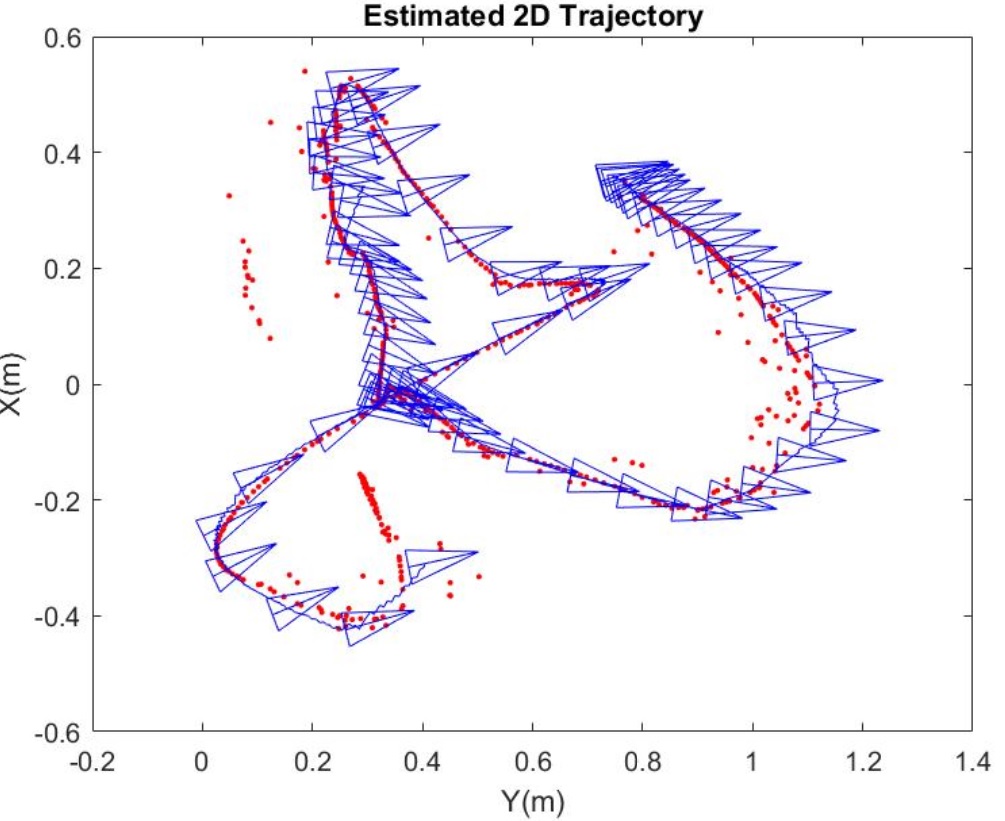,width=1\linewidth}
\caption{Estimated 2D trajectory of the robot, showing the vehicle poses (in blue) and raw measurements (in red).} \label{fig:5}
\end{figure}

\begin{figure*}[th!]
\centering
\subfloat[][Position (1st iteration)]{
\includegraphics[width=0.45\textwidth]{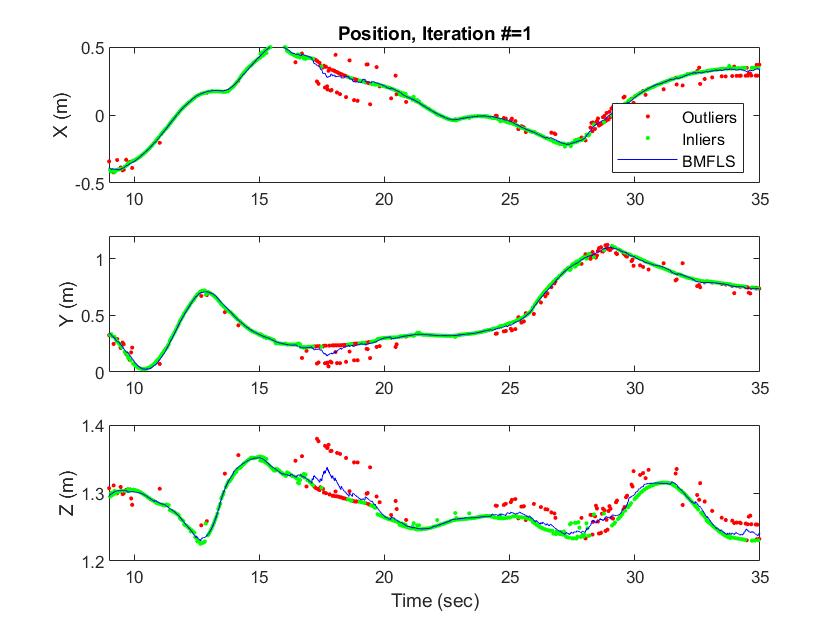}
\label{fig:subfig1}}
\qquad
\subfloat[][Position (3rd iteration)]{
\includegraphics[width=0.45\textwidth]{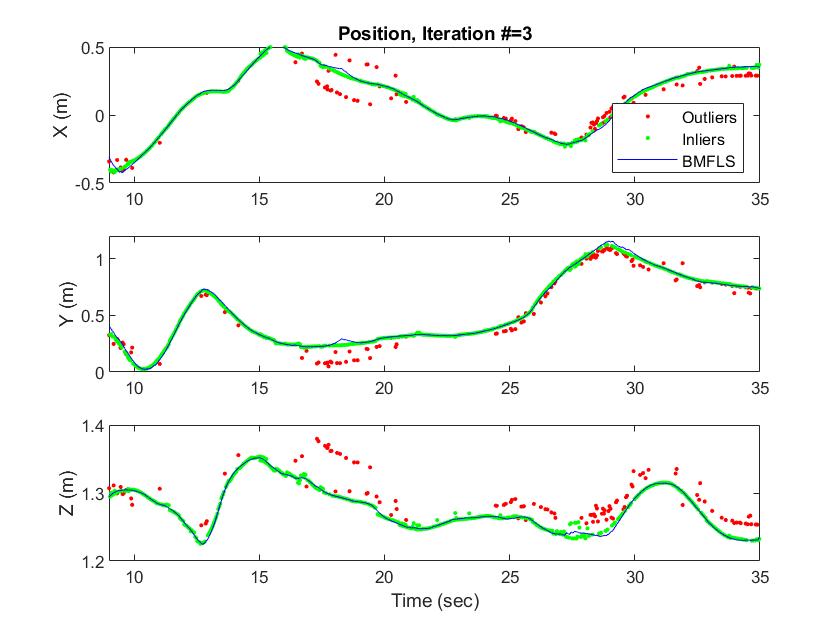}
\label{fig:subfig2}}\\
\subfloat[][Euler Angles (1st iteration)]{
\includegraphics[width=0.45\textwidth]{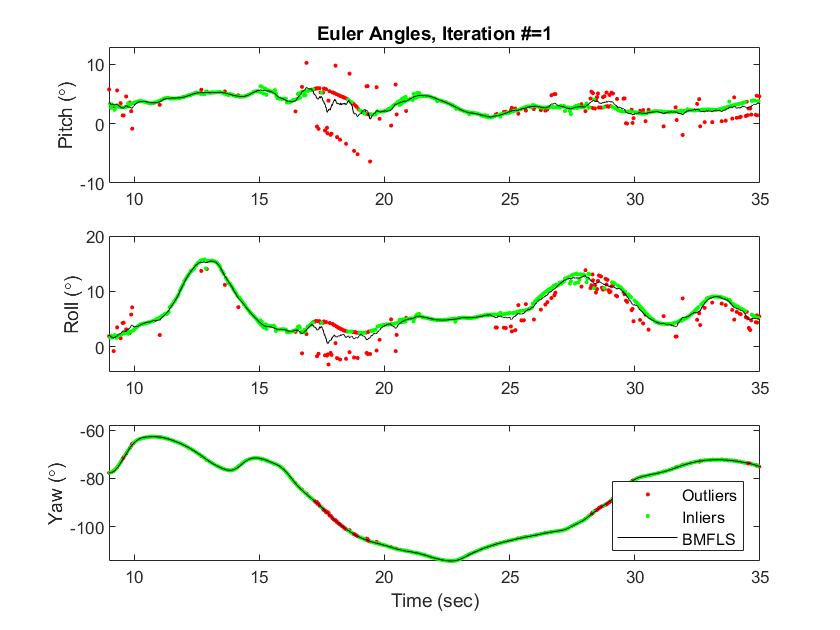}
\label{fig:subfig3}}
\qquad
\subfloat[][Euler Angles (3rd iteration)]{
\includegraphics[width=0.45\textwidth]{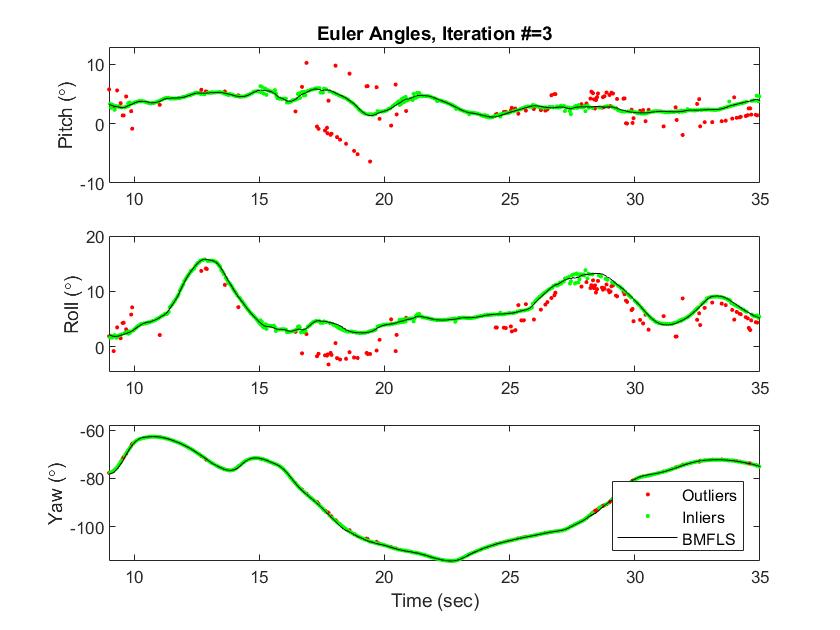}
\label{fig:subfig4}}
\caption{The first and third iteration results of the method for the position and Euler angles. It can be seen that the set of outliers gradually decreases as the iteration progresses, resulting in a better smoothing result at the 3rd iteration.}
\label{fig:4}
\end{figure*}
Figure 2 shows the visual fiducial markers (ARTag) used for the experiment. The markers are installed on the floor of the water tank facility at the UTS. The robot (shown in Figure 1) collects the data using a monocular camera and an inertial sensor. \emph{ArUco} {ROS} library is used to compute the pose measurements from the markers. Due to the lower visibility and low illumination, the markers were frequently confused with other markers, causing frequent outliers in the measurements. The robot was manually controlled to maintain a hovering position under a current disturbance source. The IMU has a sampling rate of $252$Hz, and the camera has a sampling rate of $26$Hz.

Figure \ref{fig:3} shows an example of outliers obtained from the camera/marker system. From the middle section of the data, it can be seen that there are three layers of measurements and the top two layers seem to be outliers. The ArUco marker system was likely confused with neighbouring markers, resulting in offset errors in the position. The challenge is that the inertial odometry system is also vulnerable to drift if there are few inlier measurements, which can reject all good measurements and result in filter divergences.

By applying the iterative smoothing and outlier detection method, the problem can be mitigated. Figures \ref{fig:subfig1} and \ref{fig:subfig3} show the first iteration results for the position and Euler angles in which all measurements were considered as inliers. The outliers are then declared using the smoothing results and chi-squared gating test. It can be observed that some of the inliers are wrongly classified as outliers due to the presence of multiple outliers. Figures \ref{fig:subfig2} and \ref{fig:subfig4} show the 3rd iteration results for the position and Euler angles. It can be observed that the wrongly classified inliers are now correctly declared as inliers, thanks to the improved smoothing results. Figures \ref{fig:5} shows the estimated $2$D trajectory from the proposed method, demonstrating improved accuracy due to the outlier rejections.

\section{Conclusions}\label{sec:conclusions}
We presented an iterative smoothing and outlier detection algorithm by utilising the Biswas-Mahanalobis Fixed Lag Smoother. Initially, all measurements are treated as inliers to minimise the drift in the inertial-based smoothing solution. The outliers are gradually re-classified using the chi-square gating test and the smoothed solution through successive iterations. By using a real dataset, it was shown that the algorithm converges reliably for a set of moderate-level of outliers. For a severe level of outliers, however, the filter was vulnerable to divergence. A cluster of outliers can attract the mean $\bar{x}$ and inflate covariance $S$ in its direction and away from some other observation that belongs to the pattern suggested by the majority of observation, thus yielding large values of Mahalanobis distance values for these observations  \cite{HadiAliS1992IMOi}. We are currently investigating the relaxation of the discrete indicator variable ($c\in\{0,1\}$) to a continuous one ($c\in [0,1]$) for a \emph{soft} outlier detection.

\section{Acknowledgment*}
We acknowledge Dr Andrew To and Dr Khoa Le in collecting the dataset from the robot.


\begin{thebibliography}{}

\bibitem[\protect\citeauthoryear{Agamennoni \bgroup \em et al.\egroup
  }{2011}]{AgamennoniGabriel2011AoKf}
Gabriel Agamennoni, Juan~I Nieto, and Eduardo~M Nebot.
\newblock An outlier-robust kalman filter.
\newblock In {\em 2011 IEEE International Conference on Robotics and
  Automation}, pages 1551--1558. IEEE, 2011.

\bibitem[\protect\citeauthoryear{Aggarwal}{2017}]{alma991032133873605106}
Charu~C. Aggarwal.
\newblock {\em Outlier Analysis}.
\newblock Springer International Publishing, 2nd ed. 2017. edition, 2017.

\bibitem[\protect\citeauthoryear{Bonin-Font \bgroup \em et al.\egroup
  }{2015}]{Bonin-FontFrancisco2015Vsfa}
Francisco Bonin-Font, Gabriel Oliver, Stephan Wirth, Miquel Massot, Pep
  Lluis~Negre, and Joan-Pau Beltran.
\newblock Visual sensing for autonomous underwater exploration and intervention
  tasks.
\newblock {\em Ocean engineering}, 93:25--44, 2015.

\bibitem[\protect\citeauthoryear{Chang}{2014}]{ChangGuobin2014RKfb}
Guobin Chang.
\newblock Robust kalman filtering based on mahalanobis distance as outlier
  judging criterion.
\newblock {\em Journal of geodesy}, 88(4):391--401, 2014.

\bibitem[\protect\citeauthoryear{Chavez \bgroup \em et al.\egroup
  }{2019}]{ChavezArturoGomez2019Unuv}
Arturo~Gomez Chavez, Christian~A Mueller, Tobias Doernbach, and Andreas Birk.
\newblock Underwater navigation using visual markers in the context of
  intervention missions.
\newblock {\em International journal of advanced robotic systems}, 16(2), 2019.

\bibitem[\protect\citeauthoryear{Ghorbani}{2019}]{GhorbaniHamid2019MDAI}
Hamid Ghorbani.
\newblock Mahalanobis distance and its application for detecting multivariate
  outliers.
\newblock {\em Facta universitatis. Series, mathematics and informatics}, pages
  583--, 2019.

\bibitem[\protect\citeauthoryear{Grewal and
  Andrews}{2014}]{GrewalMohinderS2014OS}
Mohinder~S Grewal and Angus~P Andrews.
\newblock Optimal smoothers.
\newblock In {\em Kalman Filtering}, Wiley - IEEE, pages 239--279. Wiley,
  Hoboken, NJ, USA, 4 edition, 2014.

\bibitem[\protect\citeauthoryear{Hadi}{1992}]{HadiAliS1992IMOi}
Ali~S Hadi.
\newblock Identifying multiple outliers in multivariate data.
\newblock {\em Journal of the Royal Statistical Society. Series B,
  Methodological}, 54(3):761--771, 1992.

\bibitem[\protect\citeauthoryear{Kautz and
  Eskofier}{2015}]{KautzThomas2015ArKf}
Thomas Kautz and Bjoern~M Eskofier.
\newblock A robust kalman framework with resampling and optimal smoothing.
\newblock {\em Sensors (Basel, Switzerland)}, 15(3):4975--4995, 2015.

\bibitem[\protect\citeauthoryear{Lee and Johnson}{2020}]{LeeKyuman2020ROFf}
Kyuman Lee and Eric~N Johnson.
\newblock Robust outlier-adaptive filtering for vision-aided inertial
  navigation.
\newblock {\em Sensors (Basel, Switzerland)}, 20(7):2036--, 2020.

\bibitem[\protect\citeauthoryear{Nam and Gon-Woo}{2020}]{NamDinhVan2020RSVI}
Dinh~Van Nam and Kim Gon-Woo.
\newblock Robust stereo visual inertial navigation system based on multi-stage
  outlier removal in dynamic environments.
\newblock {\em Sensors (Basel, Switzerland)}, 20(10):2922--, 2020.

\bibitem[\protect\citeauthoryear{Wang \bgroup \em et al.\egroup
  }{2016}]{WangRong2016CaSc}
Rong Wang, Zhi Xiong, Jianye Liu, Jianxin Xu, and Lijuan Shi.
\newblock Chi-square and sprt combined fault detection for multisensor
  navigation.
\newblock {\em IEEE transactions on aerospace and electronic systems},
  52(3):1352--1365, 2016.

\end{thebibliography}

\balance

\end{document}